\documentclass[letterpaper, 10 pt, conference]{ieeeconf}

\IEEEoverridecommandlockouts
\usepackage{cite}
\usepackage{amsmath,amssymb,amsfonts}
\usepackage{graphicx}
\usepackage{booktabs}
\usepackage{array}
\usepackage{multirow}
\usepackage{url}
\usepackage{xcolor}
\usepackage{tikz}
\usetikzlibrary{positioning, arrows.meta, calc, fit, backgrounds}
\usepackage[hidelinks]{hyperref}

\title{\LARGE \bf SAM3 Self-Distillation for Fine-Grained\\
GOOSE 2D Semantic Segmentation}

\author{Xuesong Wang$^{1}$%
\thanks{$^{1}$X. Wang conducted this work at Wayne State University,
Detroit, MI, USA. {\tt\small xswang@wayne.edu}}%
}

\begin{document}

\maketitle
\thispagestyle{empty}
\pagestyle{empty}

\begin{abstract}
We describe our 4th-place entry to the ICRA 2026 GOOSE 2D Fine-Grained
Semantic Segmentation Challenge, which reached a composite mean
Intersection-over-Union (mIoU) of \textbf{69.73\%} on the official
1{,}815-image test set. Our model adapts the image encoder of a recent
visual foundation model, Segment Anything Model~3 (SAM3), with a
lightweight decoder. Beyond this, we contribute two techniques and one
empirical finding: (i) a \emph{self-distillation} scheme that re-uses
SAM3 itself, prompted with ground-truth boxes, as a teacher on the
classes where it outperforms our own model; (ii) an \emph{image-level
multi-scale test-time augmentation} scheme that restores multi-scale
inference for a fixed-input-size model by rescaling the image rather than
the model input; and (iii) the finding that an aggressive photometric
distortion from a winning 2025 GOOSE 2D entry, transplanted onto our
pipeline, is its single largest source of improvement.
\end{abstract}

\section{Introduction}
\label{sec:intro}

Semantic segmentation, i.e., labeling every pixel of an image with the object
class it belongs to, is a core perception task for autonomous ground
robots operating off-road. The
GOOSE~\cite{mortimer2023goose} and GOOSE-Ex~\cite{hagmanns2024gooseex}
datasets provide such pixel-level annotations for unstructured outdoor
scenes captured from four ground-robot camera platforms (MuCAR-3, ALICE,
and two Spot configurations), with 64 fine-grained classes (e.g. distinct
vegetation, terrain, and vehicle types) that also roll up into 11 coarse
super-categories such as Vegetation, Terrain, and Vehicle.

The ICRA 2026 GOOSE 2D Fine-Grained Semantic Segmentation Challenge
evaluates models on a held-out 1{,}815-image test set spanning all four
platforms. Accuracy is measured by Intersection-over-Union (IoU): for a
given class, the overlap between the predicted and ground-truth pixels
divided by their union. The mean IoU (mIoU) averages this over classes.
The challenge ranks entries by a \emph{composite} score: the equal-weight
average of the fine mIoU (averaged over the 56 evaluated fine classes)
and the coarse mIoU (averaged over the 11 super-categories).

The challenge is hard for three reasons. First, the class distribution
is heavily long-tailed: a handful of head classes (e.g., forest and
low-grass) occupy more than a quarter of all training pixels, while
classes such as kick-scooter and pipe appear in only a few parts per
million. Second, many off-road categories form within-category visual
near-equivalence classes. For example, the Vegetation
supercategory contains twelve fine classes, several of which differ only
in canopy density or hardiness, and the vast majority of our baseline
model's Vegetation errors are confusions \emph{between} Vegetation
classes rather than with anything else. Third, the four camera setups
differ in resolution, exposure, and viewpoint, so a recipe that overfits
to one platform's geometry tends to lose on the others.

Our model uses the image encoder of Segment Anything Model~3
(SAM3)~\cite{carion2025sam3}, the latest in the Segment Anything family
of visual foundation models. SAM3 is \emph{promptable}: given an image
and a prompt (a point, a box, or a text phrase), it returns a mask for
the indicated object. We adapt its image encoder to off-road
segmentation by fine-tuning only its upper layers and attaching a small
decoder that produces the full per-pixel class map. Beyond this standard
recipe we make two methodological contributions:

\begin{enumerate}
\item \textbf{Self-distillation from SAM3} (Section~\ref{sec:distill}).
  We re-use SAM3 itself as a teacher for our own model, the
  same foundation model that provides our encoder. When SAM3 is given a
  ground-truth bounding box around an object, it produces an unusually
  clean mask of that object. We transfer these masks into our model as an
  extra training target, but only for the classes where we have first
  verified that SAM3's masks are genuinely better than our model's own
  predictions.
\item \textbf{Image-level multi-scale test-time augmentation}
  (Section~\ref{sec:multiscale}). Running a model on several rescaled
  copies of an image and averaging the results is a standard way to
  improve segmentation. SAM3's encoder, however, only accepts one fixed
  input size, which rules out the usual approach. We show a simple
  workaround that rescales the \emph{image} instead of the model input,
  recovering the benefit of multi-scale inference for any fixed-input
  model at no additional training cost.
\end{enumerate}

In addition, we adopt the aggressive photometric
distortion of a winning 2025 GOOSE 2D entry~\cite{kim2025goose}, which
perturbs brightness, contrast, saturation, and hue under independent
probabilities (Section~\ref{sec:photometric}); we report it as an
empirical finding rather than a novel method.

Section~\ref{sec:method} details each component,
Section~\ref{sec:experiments} reports results and a step-by-step
breakdown of where the gains come from, and
Section~\ref{sec:discussion} summarizes the approaches we tried that did
\emph{not} help.

\section{Method}
\label{sec:method}

\subsection{Backbone and Decoder}
\label{sec:backbone}

We use SAM3's pretrained image encoder as our backbone. Rather than
retrain it from scratch or freeze it entirely, we \emph{partially}
fine-tune it: only its upper layers are updated during training, while
the lower layers keep their pretrained weights. This is a deliberate
balance: the lower layers already capture generic visual
features that we do not want to disturb on a comparatively small
training set, while the upper layers adapt to the off-road domain.
Because the encoder is far larger than the decoder, we also train it
with a smaller learning rate, so the early training steps do not wash
out its pretrained knowledge. How many upper layers to unfreeze was
chosen by a short search on the validation set; the value we picked is
where unfreezing more layers stopped helping
(Section~\ref{sec:experiments}).

\begin{figure*}[t]
  \centering
  \begin{tikzpicture}[
    font=\small,
    >={Latex[length=2.2mm]},
    node distance=11mm,
    base/.style={draw, rounded corners=3pt, minimum height=13mm,
                 minimum width=26mm, align=center, inner sep=4pt, line width=0.6pt},
    io/.style   ={base, draw=black!45,  fill=black!5},
    enc/.style  ={base, draw=blue!55,   fill=blue!8},
    pyr/.style  ={base, draw=blue!45,   fill=blue!4},
    dec/.style  ={base, draw=teal!60,   fill=teal!9},
    teach/.style={base, draw=orange!80, fill=orange!10},
    loss/.style ={base, draw=red!55,    fill=red!7},
    fwd/.style  ={->, line width=0.8pt, draw=black!65},
    aux/.style  ={->, line width=0.8pt, draw=orange!75, dashed},
  ]
    \node[io] (img) {Input image\\[-1pt]{\scriptsize$1008\times1008$}};
    \node[enc, right=of img] (enc) {SAM3 encoder\\[-1pt]{\scriptsize upper 12 blocks}\\[-2pt]{\scriptsize fine-tuned, rest frozen}};
    \node[pyr, right=of enc] (pyr) {Feature pyramid\\[-1pt]{\scriptsize 4 levels}};
    \node[dec, right=of pyr] (dec) {FPN-style\\[-1pt]fusion decoder};
    \node[io, right=of dec] (out) {Per-pixel\\[-1pt]class map};
    \foreach \a/\b in {img/enc,enc/pyr,pyr/dec,dec/out} \draw[fwd] (\a)--(\b);
    \node[io, below=16mm of img] (gt) {GT labels\\[-1pt]$+$ boxes};
    \node[teach, below=16mm of enc] (teacher) {SAM3 teacher\\[-1pt]{\scriptsize oracle-box prompt}};
    \node[loss, below=16mm of dec, minimum width=40mm] (loss) {Training loss\\[-1pt]{\scriptsize$\mathcal{L}_{\mathrm{CE}}$ (56 cls)\,$+\,\lambda\,\mathcal{L}_{\mathrm{teach}}$ (22 cls)}};
    \draw[fwd] (gt) -- (teacher);
    \draw[aux] (teacher) -- (loss) node[midway, above, font=\scriptsize, text=black!60] {soft masks};
    \draw[fwd] (gt.south) -- ++(0,-4mm) -| (loss.south);
    \draw[fwd] (out.south) |- (loss.east);
    \begin{scope}[on background layer]
      \node[draw=black!30, dotted, line width=0.8pt, rounded corners=5pt,
            fill=black!3, fit=(gt)(teacher)(loss), inner sep=5mm] (grp) {};
    \end{scope}
    \node[anchor=north west, font=\scriptsize\itshape, text=black!55]
      at ([shift={(1mm,-1mm)}]grp.north west) {training only};
  \end{tikzpicture}
  \caption{Model architecture. The deployed model is the top row: a
    partially fine-tuned SAM3 image encoder produces a four-level feature
    pyramid that a lightweight FPN-style decoder fuses into a per-pixel
    class map. The bottom row is used only during training: SAM3 itself,
    prompted with ground-truth boxes (oracle-box), provides soft masks
    that supervise the student on 22 selected classes through
    $\mathcal{L}_{\text{teach}}$, alongside the standard cross-entropy
    loss on all 56 classes.}
  \label{fig:arch}
\end{figure*}
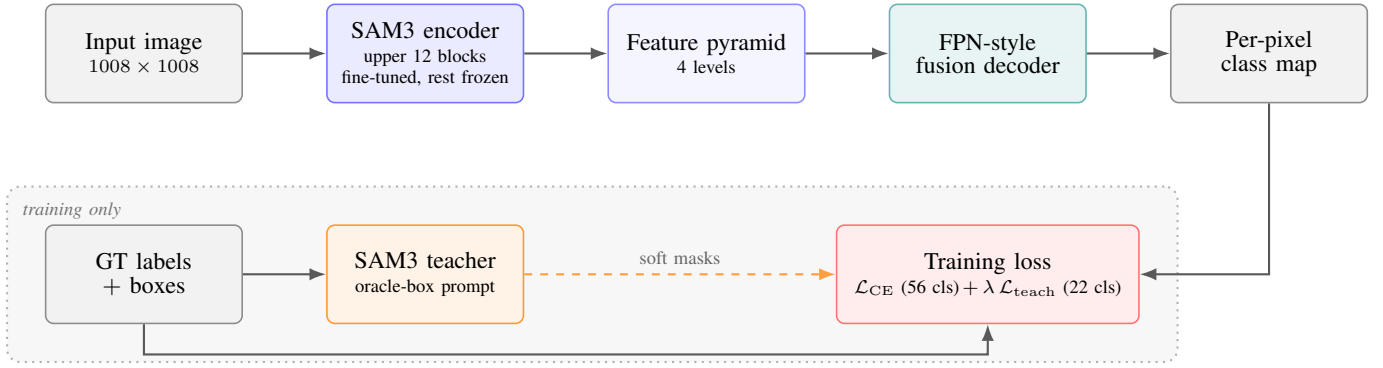

The encoder emits features at several resolutions. On top of it we
attach a lightweight decoder in the style of a Feature Pyramid Network
(FPN)~\cite{lin2017fpn}: the multi-resolution features are brought to a
common resolution, combined, and passed through a small convolutional
head that outputs the final per-pixel class map (Figure~\ref{fig:arch}).
We keep this decoder intentionally simple. We also tried a heavier, more sophisticated
decoder (UPerNet~\cite{xiao2018upernet}) as a comparison
(Section~\ref{sec:discussion}); it scored about half a point of
composite mIoU lower, so we kept the simpler design.

\subsection{SAM3 Oracle-Box Self-Distillation}
\label{sec:distill}

\paragraph{Why SAM3 can teach our model}
SAM3 was pretrained on a far larger and more varied collection of
images than the GOOSE training set, so it has a strong general sense of
object shape. The question is how to extract that knowledge for our
specific classes. Prompting SAM3 with only a class name (its text mode)
does not work on off-road scenes: in a quick check, prompting for
``water'' returned almost every textured region as water, whether or not
it actually was. The fix is to also give SAM3 the ground-truth bounding
box of each object as a prompt. The box confines SAM3 to the right
region, and within that region its mask is usually sharper and more
complete than our model's. We call this \emph{oracle-box} prompting,
because the box is taken from the ground-truth annotation.

\paragraph{Which classes to teach}
SAM3 is not better than our model on every class. On large background
regions (e.g. walls, rock) and thin connected structures (e.g. fences,
guard rails) its mask tends to spill outside the box, making it worse,
not better. We therefore run a one-time check before training: for each
class we compare SAM3's oracle-box mask against our model's own
predictions on held-out images, and keep a class only when SAM3 is
clearly better and we have enough labelled pixels to trust the
comparison. This leaves 22 classes (out of the 56 evaluated),
overwhelmingly compact, well-defined objects, exactly the
kind of object SAM3's pretraining has seen in abundance: moss, truck,
hedge, leaves, bush, tree-crown, container, misc-sign, traffic-cone,
water, motorcycle, high-grass, kick-scooter, debris, barrier-tape,
traffic-sign, caravan, obstacle, trailer, bicycle, traffic-light, and
bus.

\paragraph{How the teaching signal is added}
During training our model is supervised in the usual way,
by a standard per-pixel classification loss (cross-entropy, CE) against
the ground-truth labels across all 56 classes. On top of this, for the
22 selected classes only, we add a second loss term that asks our
model's predicted mask to agree with SAM3's mask. Writing
$\mathcal{L}_{\text{CE}}$ for the standard term and
$\mathcal{L}_{\text{teach}}$ for the SAM3-agreement term, the model
minimizes
\[
\mathcal{L} \;=\; \mathcal{L}_{\text{CE}} \;+\; \lambda \cdot \mathcal{L}_{\text{teach}},
\]
where $\lambda$ is a small weight that keeps the SAM3 signal a refinement
rather than the dominant objective. Because SAM3 is far too slow to query
inside the training loop, we compute its masks once for every training
image ahead of time and store them. When an image is randomly cropped or
flipped during training, we apply the identical geometric transform to
the stored SAM3 mask so the two stay aligned; color augmentation is
applied to the image only.

\subsection{Aggressive Color Augmentation}
\label{sec:photometric}

The off-road scenes in GOOSE span seasons (winter snow, summer
vegetation), weather (sunny, overcast, rain), and different camera
exposure settings. A model that leans on color as a shortcut for
recognizing classes is therefore fragile: the moment lighting or season
changes, its cue disappears. The standard remedy is color augmentation:
during training the image is randomly perturbed in
brightness, contrast, saturation, and hue, so the model is pushed to
recognize classes by shape and texture rather than color alone.

We adopt the aggressive photometric distortion introduced by Kim et
al.~\cite{kim2025goose} for the 2025 GOOSE 2D challenge, where it was a
central ingredient of a winning entry. Whereas the conventional
recipe~\cite{liu2016ssd} applies all four perturbations together under a
single random switch, this version gives each of the four its own
independent switch, so any subset may be applied to a given training
image. This produces a much richer mix of training images: most images
receive one or two perturbations rather than either all four or none.
Table~\ref{tab:photometric} lists the perturbation strengths, which
follow Kim et al.~\cite{kim2025goose}. Our contribution here is not the augmentation itself
but the finding that, transplanted onto our SAM3 distillation pipeline,
it is by a wide margin the single largest source of improvement.

\begin{table}[t]
  \centering
  \caption{Color-perturbation strengths; each is applied independently
    with probability $0.5$ (saturation and hue act in HSV space).}
  \label{tab:photometric}
  \small
  \begin{tabular}{lll}
    \toprule
    Perturbation & Type & Range \\
    \midrule
    Brightness & additive   & $\pm 40/255$ \\
    Contrast   & multiplicative & $[0.7, 1.3]$ \\
    Saturation & multiplicative & $[0.7, 1.3]$ \\
    Hue        & additive     & $\pm 0.1$    \\
    \bottomrule
  \end{tabular}
\end{table}

\subsection{Image-Level Multi-Scale Sliding-Window TTA}
\label{sec:multiscale}

Multi-scale test-time augmentation (TTA) is a standard inference-time
technique: run the model on several rescaled versions of the input
image, then average the results. Seeing each object at several sizes
makes the prediction more robust and typically gives a free fraction of
a point of mIoU. The catch in our setting is that SAM3's encoder was
pretrained at a single fixed input size and rejects images of any other
size, so we cannot simply feed it rescaled images the way the standard
recipe does.

\paragraph{The image-level workaround.}
We move the scaling \emph{outside} the model. Because the test images
are larger than the model's fixed input, we already process them by
\emph{tiling} (sliding-window inference), and multi-scale augmentation
fits naturally around this loop. For each scale factor in a small
predefined set (here, $\{0.75, 1.0, 1.25\}$):

\begin{enumerate}
\item \textbf{Rescale} the whole test image to the chosen scale.
\item \textbf{Tile} the rescaled image with overlapping native-size
  ($1008\times1008$) windows and score each window independently; the
  model always sees its fixed input size, only the image being windowed
  changes.
\item \textbf{Fuse} the overlapping window scores into one
  full-resolution score map, weighting each window by a raised-cosine
  (Hann) profile so tile edges are down-weighted and the seams between
  tiles are smoothed.
\item \textbf{Resize} that score map back to the original image
  resolution and accumulate it across scales.
\end{enumerate}

\noindent After all scales have been processed, the final prediction is
the class with the highest accumulated score at each pixel.

This idea combines cleanly with the other common test-time trick of
also running each window on its mirror image and averaging
(``horizontal-flip'' augmentation); we use both. It is also general: any
model that is locked to a single input size can recover the benefit of
multi-scale inference this way, without retraining. The cost grows
linearly with the number of scales; with three scales, inference takes
roughly three times as long as a single pass.

\begin{figure*}[t]
  \centering
  \includegraphics[width=\textwidth]{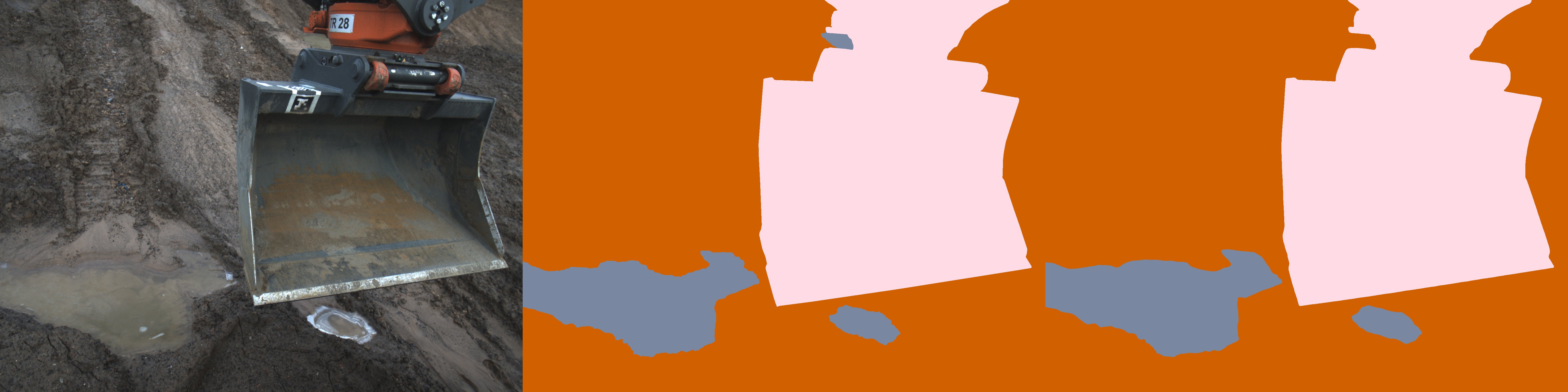}\\
  \vspace{1mm}
  \includegraphics[width=\textwidth]{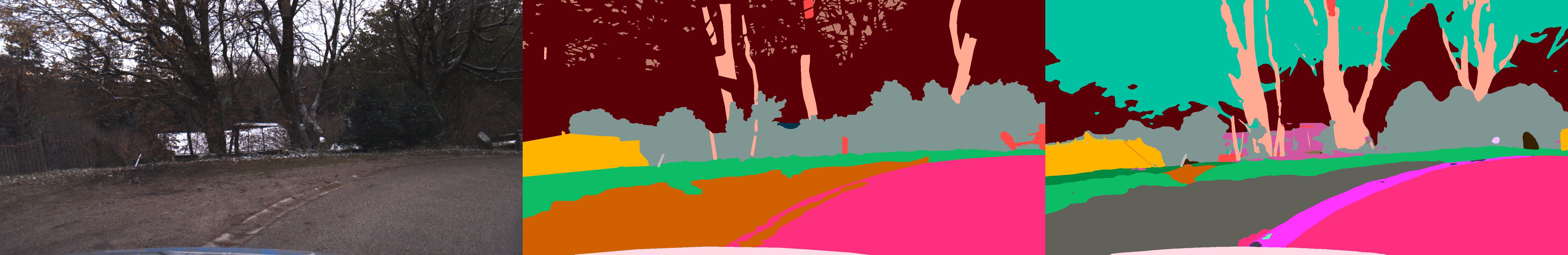}
  \caption{Qualitative results on the validation set. Each row shows the
    input image (left), the ground-truth labeling colored by class
    (center), and our model's prediction (right). The top row is one of
    the strongest scenes (composite mIoU 96.6\%), with a clean separation
    of sky, vegetation, road, and vehicle. The bottom row is one of the
    weakest (composite mIoU 32.8\%): a cluttered winter scene where the
    model confuses visually similar vegetation classes such as bush,
    low-grass, and leaves, which account for most of our remaining
    errors.}
  \label{fig:qualitative}
\end{figure*}

\section{Experiments}
\label{sec:experiments}

\subsection{Setup}

\paragraph{Data.}
We train on the union of the GOOSE 2D and GOOSE-Ex 2D \emph{train}
splits (11{,}234 standard color, i.e. red-green-blue (RGB), images;
the optional near-infrared channel provided by GOOSE was not used in
this entry) and validate on the union
of the two \emph{val} splits (1{,}369 images). The final test set is
the official 1{,}815-image GOOSE 2D Final Testing split, covering all
four robotic platforms (MuCAR-3, ALICE, Spot v1, Spot v2).

\paragraph{Implementation details.}
SAM3's encoder takes a fixed $1008 \times 1008$ RGB image and produces
features at four resolutions, which our FPN-style decoder fuses. We
fine-tune the last 12 transformer blocks of the encoder together with
its feature-fusion module, leaving the earlier blocks frozen; we chose
12 after a short search over $\{4, 8, 12, 16\}$, where unfreezing 16
blocks did not improve over 12. Training uses the AdamW
optimizer~\cite{loshchilov2019adamw} with a base learning rate of
$3 \times 10^{-5}$, reduced by a factor of four on the encoder, weight
decay $0.01$, a polynomial learning-rate decay, batch size $2$, and
mixed-precision arithmetic. The teaching-signal weight
(Section~\ref{sec:distill}) is $\lambda = 0.3$. We train for up to 25
epochs over the data, selecting the model with the best validation
composite mIoU; the model is initialised from our earlier
self-distillation model (the ``$+$~SAM3 self-distillation'' configuration
of Table~\ref{tab:ablation}), i.e.\ the same recipe \emph{without} the
color augmentation, which in turn was fine-tuned from a plain
cross-entropy model. A full run takes about 8.5 hours on a single NVIDIA RTX~4090 GPU with 24~GB of memory. At
inference we tile each image with overlapping $1008 \times 1008$ windows
at an 840-pixel stride ($\sim$17\% overlap), fused with Hann weighting as
described in Section~\ref{sec:multiscale}.
On top of this we apply both horizontal-flip and image-level multi-scale
augmentation at scales $\{0.75, 1.0, 1.25\}$, averaging the model's raw per-class
scores across all the variants before choosing each pixel's label.

\subsection{Main Results}

Table~\ref{tab:ablation} traces our submissions to the public test
leaderboard and the improvement each contribution adds, all measured on
the same 1{,}815-image test set under directly comparable settings.

\begin{table}[t]
  \centering
  \caption{Effect of each component on the public test leaderboard
    (mIoU \%): non-indented rows are cumulative training milestones, and
    each indented row adds test-time augmentation to the milestone above
    it (last row: our final entry).}
  \label{tab:ablation}
  \small
  \setlength{\tabcolsep}{4pt}
  \begin{tabular}{lccc}
    \toprule
    Configuration & composite & fine & coarse \\
    \midrule
    Encoder + decoder (CE loss) & 67.71 & 62.33 & 73.10 \\
    + SAM3 self-distillation                & 68.38 & 62.15 & 74.61 \\
    \quad + flip augmentation               & 68.51 & 62.35 & 74.67 \\
    + color augmentation                   & 69.24 & 62.80 & 75.68 \\
    \quad + flip augmentation               & 69.39 & 63.04 & 75.73 \\
    \quad + flip \& multi-scale (\textbf{final}) & \textbf{69.73} & \textbf{63.47} & \textbf{75.99} \\
    \bottomrule
  \end{tabular}
\end{table}

The largest single gain comes from color augmentation ($+0.86$ over the
self-distillation model). Flip augmentation is a small but reliable
$+0.13$ to $+0.15$ on either model, and image-level multi-scale
augmentation on top of that adds a further $+0.34$.

\subsection{Per-Class Analysis}

Table~\ref{tab:perclass} reports per-class IoU on a representative subset
of classes at three stages of our pipeline. Two patterns stand out.
First, color augmentation rescues rare classes that the earlier model
had largely missed; the single largest class-level gain we saw in any
experiment was on kick-scooter ($+21.75$~IoU). Second, the test-time
augmentation stack (flip and multi-scale) most helps medium-rare,
well-defined object classes such as traffic-cone, boom-barrier,
traffic-sign, and heavy-machinery, several of which color
augmentation alone had left flat or even hurt (traffic-cone, for
instance, dropped under color augmentation and was then partly
recovered). We suspect the multi-scale component drives much of this,
with the smaller $0.75\times$ view supplying whole-object context and the
$1.25\times$ view sharpening fine detail, though our experiments isolate
multi-scale only at the aggregate level ($+0.34$ composite, Table~\ref{tab:ablation}).

\begin{table}[t]
  \centering
  \caption{Per-class IoU (\%) on the public test set at three stages:
    \textbf{Base} (encoder, decoder, and SAM3 self-distillation),
    \textbf{+Color} (adds the color augmentation), and \textbf{Final}
    (adds flip and image-level multi-scale test-time augmentation).}
  \label{tab:perclass}
  \small
  \setlength{\tabcolsep}{4pt}
  \begin{tabular}{lrrr}
    \toprule
    Class & Base & +Color & Final \\
    \midrule
    \multicolumn{4}{l}{\textit{Improved by color augmentation (Base $\rightarrow$ +Color)}} \\
    \quad kick-scooter    &  7.90 & 29.65 & 30.02 \\
    \quad debris          & 14.45 & 26.92 & 28.63 \\
    \quad bridge          &  6.67 & 17.04 & 15.96 \\
    \quad truck           & 55.96 & 65.46 & 67.89 \\
    \quad water           & 44.09 & 46.65 & 47.27 \\
    \midrule
    \multicolumn{4}{l}{\textit{Improved by test-time augmentation (+Color $\rightarrow$ Final)}} \\
    \quad traffic-cone    & 73.76 & 62.16 & 67.93 \\
    \quad boom-barrier    & 63.23 & 62.01 & 65.74 \\
    \quad traffic-sign    & 72.25 & 72.71 & 75.12 \\
    \quad heavy-machinery & 63.45 & 66.13 & 67.96 \\
    \quad tree-crown      & 55.09 & 54.44 & 56.26 \\
    \midrule
    \multicolumn{4}{l}{\textit{Persistent failures}} \\
    \quad moss            &  9.93 &  0.01 &  0.00 \\
    \quad tree-root       &  0.15 &  0.00 &  0.01 \\
    \bottomrule
  \end{tabular}
\end{table}

The clearest persistent failure is moss: the earlier model scored it
modestly, and the color-augmented model collapsed it to near zero, which
multi-scale augmentation does not recover. Moss is one of the rarest
classes in the training set and looks much like low-grass and forest;
this kind of within-vegetation confusion is visible in the failure case
of Figure~\ref{fig:qualitative}. An
ensemble that used our earlier model for moss and the final model
elsewhere would plausibly recover it, but our attempts to combine the two
models by simply averaging their scores gave almost nothing, because such
averaging follows whichever model is more \emph{confident} rather than
whichever is more \emph{accurate}. The per-super-category results in
Table~\ref{tab:percoarse} tell the same story at a coarser level: Sky,
Vegetation, and Vehicle are very strong, while Water and Animal remain
weak ($<$50~IoU). Both weak categories consist of a single rare class
and would likely benefit most from additional pretraining on related
off-road or driving datasets.

\begin{table}[t]
  \centering
  \caption{Per-super-category (coarse) mIoU (\%) for our final entry on
    the public test set.}
  \label{tab:percoarse}
  \small
  \setlength{\tabcolsep}{4pt}
  \begin{tabular}{lr lr}
    \toprule
    Category & IoU & Category & IoU \\
    \midrule
    Sky          & 97.47 & Construction & 79.79 \\
    Vegetation   & 94.01 & Road         & 72.21 \\
    Human        & 91.58 & Object       & 56.59 \\
    Vehicle      & 90.94 & Water        & 47.27 \\
    Sign         & 88.82 & Animal       & 28.44 \\
    Terrain      & 88.76 &              &       \\
    \bottomrule
  \end{tabular}
\end{table}

\section{Discussion: What Did Not Help}
\label{sec:discussion}

For completeness, we briefly record the approaches we tried that did not
improve our score and were therefore not part of the final entry. We
believe these negative results are as useful to future participants as
the positive ones.

\textbf{Alternative backbones.} Before settling on SAM3 we evaluated
frozen-encoder baselines using DINOv2~\cite{oquab2023dinov2},
I-JEPA~\cite{assran2023ijepa}, InternImage~\cite{wang2023internimage},
and a Mask2Former~\cite{cheng2022mask2former} head, all on the same data
and loss. On our validation split their best composite mIoU ranged from
30.1\% (I-JEPA) to 53.3\% (Mask2Former), well short of a comparably frozen
SAM3 encoder (60.2\%), so we did not pursue them. Our strongest pre-SAM3
model was a fully-trained SegFormer~\cite{xie2021segformer} (the
challenge's baseline architecture), which reached 50.0\% composite mIoU on
the public development leaderboard; frozen and then partially fine-tuned
SAM3 surpassed it (50.4\% and 53.7\%), which is why our final system is built
on SAM3.

\textbf{A heavier decoder.} We replaced our simple decoder with the more
elaborate UPerNet decoder~\cite{xiao2018upernet}. It scored about half a
point lower overall, even though it won on roughly a third of the
individual classes: it helped on large background regions (walls, gravel,
bridges) but blurred away small or rare objects.

\textbf{Random rescaling during training.} Adding random image rescaling
to the color-augmentation recipe did not beat the color-only model on
the validation set, though it produced a noticeably different model that
we kept as a possible ensemble partner for future work.

\textbf{A long-tail loss (Balancing Logit Variation).} We added a
published training technique~\cite{wang2023blv} designed to help rare
classes by injecting class-dependent noise into the model's predictions
during training. It scored
slightly below our best model on the validation set, so we did not
pursue it. Our implementation is available on request.

\section{Conclusion}
\label{sec:conclusion}

We described our 4th-place entry to the ICRA 2026 GOOSE 2D Challenge
(composite mIoU 69.73\%). The model adapts SAM3's pretrained image
encoder with a lightweight decoder, and contributes two techniques: (i)
using SAM3 itself as a teacher to refine our model on the classes where
the foundation model is verifiably better, and (ii) a multi-scale
inference scheme that restores the benefit of multi-scale test-time
augmentation for a model locked to a single input size, by rescaling the
image rather than the model input. We also adopt the aggressive
photometric distortion of a winning 2025 GOOSE 2D entry which proved to be the single biggest improvement in our pipeline; the
cheapest improvement came from the multi-scale inference, which adds
accuracy without any extra training.

\section*{Acknowledgements}

We thank the GOOSE Challenge organizers for hosting the ICRA 2026 challenge.

The author used artificial-intelligence (AI) tools, specifically large
language models, to help edit this report and to assist with
software development during the competition. All experiments, results,
and conclusions were produced and verified by the author, who takes full
responsibility for the content.

\bibliographystyle{IEEEtran}
\bibliography{references}

@inproceedings{mortimer2023goose,
  author = {Peter Mortimer and Raphael Hagmanns and Miguel Granero
            and Thorsten Luettel and Janko Petereit and Hans-Joachim Wuensche},
  title = {The GOOSE Dataset for Perception in Unstructured Environments},
  url={https://arxiv.org/abs/2310.16788},
  booktitle={Proceedings of the IEEE International Conference on Robotics and Automation (ICRA)},
  year = {2024}
}

@inproceedings{hagmanns2024gooseex,
  author = {Raphael Hagmanns and Peter Mortimer and Miguel Granero
            and Thorsten Luettel and Janko Petereit},
  title = {Excavating in the Wild: The GOOSE-Ex Dataset for Semantic Segmentation},
  url={https://arxiv.org/abs/2409.18788},
  booktitle={Proceedings of the IEEE International Conference on Robotics and Automation (ICRA)},
  year = {2025}
}

@inproceedings{carion2025sam3,
title={{SAM} 3: Segment Anything with Concepts},
author={Nicolas Carion and Laura Gustafson and Yuan-Ting Hu and Shoubhik Debnath and Ronghang Hu and Didac Suris Coll-Vinent and Chaitanya Ryali and Kalyan Vasudev Alwala and Haitham Khedr and Andrew Huang and Jie Lei and Tengyu Ma and Baishan Guo and Arpit Kalla and Markus Marks and Joseph Greer and Meng Wang and Peize Sun and Roman R{\"a}dle and Triantafyllos Afouras and Effrosyni Mavroudi and Katherine Xu and Tsung-Han Wu and Yu Zhou and Liliane Momeni and RISHI HAZRA and Shuangrui Ding and Sagar Vaze and Francois Porcher and Feng Li and Siyuan Li and Aishwarya Kamath and Ho Kei Cheng and Piotr Dollar and Nikhila Ravi and Kate Saenko and Pengchuan Zhang and Christoph Feichtenhofer},
booktitle={The Fourteenth International Conference on Learning Representations},
year={2026},
url={https://openreview.net/forum?id=r35clVtGzw}
}

@inproceedings{lin2017fpn,
  title={Feature pyramid networks for object detection},
  author={Lin, Tsung-Yi and Doll{\'a}r, Piotr and Girshick, Ross and He, Kaiming and Hariharan, Bharath and Belongie, Serge},
  booktitle={Proceedings of the IEEE conference on computer vision and pattern recognition},
  pages={2117--2125},
  year={2017}
}

@inproceedings{xiao2018upernet,
  title={Unified perceptual parsing for scene understanding},
  author={Xiao, Tete and Liu, Yingcheng and Zhou, Bolei and Jiang, Yuning and Sun, Jian},
  booktitle={Proceedings of the European conference on computer vision (ECCV)},
  pages={418--434},
  year={2018}
}

@inproceedings{liu2016ssd,
  title={Ssd: Single shot multibox detector},
  author={Liu, Wei and Anguelov, Dragomir and Erhan, Dumitru and Szegedy, Christian and Reed, Scott and Fu, Cheng-Yang and Berg, Alexander C},
  booktitle={European conference on computer vision},
  pages={21--37},
  year={2016},
  organization={Springer}
}

@inproceedings{cheng2022mask2former,
  title={Masked-attention mask transformer for universal image segmentation},
  author={Cheng, Bowen and Misra, Ishan and Schwing, Alexander G and Kirillov, Alexander and Girdhar, Rohit},
  booktitle={Proceedings of the IEEE/CVF conference on computer vision and pattern recognition},
  pages={1290--1299},
  year={2022}
}

@inproceedings{loshchilov2019adamw,
title={Decoupled Weight Decay Regularization},
author={Ilya Loshchilov and Frank Hutter},
booktitle={International Conference on Learning Representations},
year={2019},
url={https://openreview.net/forum?id=Bkg6RiCqY7},
}

@article{xie2021segformer,
  title={SegFormer: Simple and efficient design for semantic segmentation with transformers},
  author={Xie, Enze and Wang, Wenhai and Yu, Zhiding and Anandkumar, Anima and Alvarez, Jose M and Luo, Ping},
  journal={Advances in neural information processing systems},
  volume={34},
  pages={12077--12090},
  year={2021}
}

@article{oquab2023dinov2,
title={{DINO}v2: Learning Robust Visual Features without Supervision},
author={Maxime Oquab and Timoth{\'e}e Darcet and Th{\'e}o Moutakanni and Huy V. Vo and Marc Szafraniec and Vasil Khalidov and Pierre Fernandez and Daniel HAZIZA and Francisco Massa and Alaaeldin El-Nouby and Mido Assran and Nicolas Ballas and Wojciech Galuba and Russell Howes and Po-Yao Huang and Shang-Wen Li and Ishan Misra and Michael Rabbat and Vasu Sharma and Gabriel Synnaeve and Hu Xu and Herve Jegou and Julien Mairal and Patrick Labatut and Armand Joulin and Piotr Bojanowski},
journal={Transactions on Machine Learning Research},
issn={2835-8856},
year={2024},
url={https://openreview.net/forum?id=a68SUt6zFt},
note={Featured Certification}
}

@inproceedings{assran2023ijepa,
  title={Self-supervised learning from images with a joint-embedding predictive architecture},
  author={Assran, Mahmoud and Duval, Quentin and Misra, Ishan and Bojanowski, Piotr and Vincent, Pascal and Rabbat, Michael and LeCun, Yann and Ballas, Nicolas},
  booktitle={Proceedings of the IEEE/CVF conference on computer vision and pattern recognition},
  pages={15619--15629},
  year={2023}
}

@inproceedings{wang2023internimage,
  title={Internimage: Exploring large-scale vision foundation models with deformable convolutions},
  author={Wang, Wenhai and Dai, Jifeng and Chen, Zhe and Huang, Zhenhang and Li, Zhiqi and Zhu, Xizhou and Hu, Xiaowei and Lu, Tong and Lu, Lewei and Li, Hongsheng and others},
  booktitle={Proceedings of the IEEE/CVF conference on computer vision and pattern recognition},
  pages={14408--14419},
  year={2023}
}

@inproceedings{wang2023blv,
  title={Balancing logit variation for long-tailed semantic segmentation},
  author={Wang, Yuchao and Fei, Jingjing and Wang, Haochen and Li, Wei and Bao, Tianpeng and Wu, Liwei and Zhao, Rui and Shen, Yujun},
  booktitle={Proceedings of the IEEE/CVF conference on computer vision and pattern recognition},
  pages={19561--19573},
  year={2023}
}

@article{kim2025goose,
  title={Technical Report for ICRA 2025 GOOSE 2D Semantic Segmentation Challenge: Boosting Off-Road Segmentation via Photometric Distortion and Exponential Moving Average},
  author={Kim, Wonjune and Lee, Lae-kyoung and An, Su-Yong},
  journal={arXiv preprint arXiv:2505.11769},
  year={2025}
}

\end{document}